\documentclass[sn-mathphys,Numbered]{sn-jnl}

\usepackage{graphicx}%
\usepackage{multirow}%
\usepackage{amsmath,amssymb,amsfonts}%
\usepackage{mathrsfs}%
\usepackage{xcolor}%
\usepackage{textcomp}%
\usepackage{manyfoot}%
\usepackage{booktabs}%
\usepackage{algorithm}%
\usepackage{algorithmicx}%
\usepackage{algpseudocode}%
\usepackage{listings}%
\usepackage{hyperref}       
\usepackage{url}            
\usepackage{booktabs}       
\usepackage{amsfonts}       
\usepackage{nicefrac}       
\usepackage{microtype}      
\usepackage{xcolor}         
\usepackage{float}
\usepackage{graphicx}
\usepackage{amsthm}
\usepackage{subcaption}
\usepackage{graphicx}
\usepackage{color}
\usepackage{algorithm}
\usepackage{caption}
\usepackage{subcaption}
\usepackage{multicol}
\usepackage{comment}
\usepackage{tikz}
\usetikzlibrary{positioning}
\usepackage{graphicx}
\usepackage{xspace}
\usepackage{bm}
\usepackage{adjustbox}
\usepackage{tabularx}
\usepackage{color}
\usepackage{algorithm}
\usepackage{listings}
\makeatletter
\DeclareRobustCommand\onedot{\futurelet\@let@token\@onedot}
\def\@onedot{\ifx\@let@token.\else.\null\fi\xspace}

\def\eg{\emph{e.g}\onedot} 
\def\ie{\emph{i.e}\onedot}

\makeatother

\theoremstyle{thmstyleone}%
\theoremstyle{thmstyletwo}%

\theoremstyle{thmstylethree}%

\begin{document}

\title[Article Title]{Learning a Consensus Sub-Network with Polarization Regularization and One Pass Training}


\author{\fnm{Xiaoying} \sur{Zhi}}\email{xiaoying.zhi@jpmchase.com}

\author{\fnm{Varun} \sur{Babbar}}\email{varun.babbar@jpmchase.com}

\author{\fnm{Rundong} \sur{Liu}}\email{eric.liu@jpmchase.com}

\author{\fnm{Pheobe} \sur{Sun}}\email{pheobe.sun@jpmchase.com}

\author{\fnm{Fran} \sur{Silavong}}\email{fran.silavong@jpmchase.com}

\author{\fnm{Ruibo} \sur{Shi}}\email{ruibo.shi@jpmchase.com}

\author{\fnm{Sean} \sur{Moran}}\email{sean.j.moran@jpmchase.com}

\affil{\orgname{JPMorgan Chase}, \orgaddress{\street{25 Bank Street}, \city{London}, \postcode{E145JP}, \country{UK}}}

\abstract{The subject of green AI has been gaining attention within the deep learning community given the recent trend of ever larger and more complex neural network models. Existing solutions for reducing the computational load of training at inference time usually involve pruning the network parameters. Pruning schemes often create extra overhead either by iterative training and fine-tuning for static pruning or repeated computation of a dynamic pruning graph. We propose a new parameter pruning strategy for learning a lighter-weight sub-network that minimizes the energy cost while maintaining comparable performance to the fully parameterised network on given downstream tasks. Our proposed pruning scheme is green-oriented, as it only requires a one-off training to discover the optimal static sub-networks by dynamic pruning methods. The pruning scheme consists of a binary gating module and a polarizing loss function to uncover sub-networks with user-defined sparsity. Our method enables pruning and training simultaneously, which saves energy in both the training and inference phases and avoids extra computational overhead from gating modules at inference time. Our results on CIFAR-10, CIFAR-100, and Tiny Imagenet suggest that our scheme can remove ${\approx{}50}\%$ of connections in deep networks with $ \leq 1\%$ reduction in classification accuracy. Compared to other related pruning methods, our method demonstrates a lower drop in accuracy for equivalent reductions in computational cost.}




\maketitle

\section{Introduction}\label{sec1}

Large, sparse and over-parameterised models bring state-of-the-art (SOTA) performance on many tasks but require significantly more computational power and therefore energy compared to conventional machine learning models \citep{greenai_schwartz_2020}. For example, the vision transformer model (ViT-L16) with 307M parameters can achieve 99.42\% accuracy on the CIFAR-10 dataset and 87.76\% on the ImageNet dataset \citep{ViT_dosovitskiy_2020}. The training of ViT-L16 model requires 680 TPUv3-core-days \footnote{Multiplication of the number of used TPUv3 cores and the training time in days} and 3672kWh energy consumption, equivalent to 32.5\% of the annual energy consumption of an average US household \citep{
ViT_dosovitskiy_2020,tpu_jouppi_2020,energy_eia_2021}.

Network pruning is a promising research direction that helps achieve greener AI models, based on the assumption that we can safely remove parameters from over-parameterized networks, without significantly degrading the network performance~\cite{predictparams_denil_2013}. There are two common types of network pruning methods - \emph{static} and \emph{dynamic}. Static network pruning generates a unified sub-network for all data, while dynamic pruning computes different sub-networks for each data sample. Static network pruning often requires a pre-defined neuron importance measure which determines which trained neurons should be pruned \cite{dgate_shafiee_2018,thinnet_luo_2019,transformerzip_cheong_2019,oneshotpruning_zhang_2019,sniptransformerpruning_lin_2020}. Further fine-tuning or regrowing of the selected sub-network are often involved after training, which can potentially lead to further improvement in performance \cite{threephasepruning_han_2015,vitpruning_zhu_2021,chex_hou_2022}. Dynamic pruning, on the other hand, applies a parameterized and learnable gating function that computes the neuron importance on the fly, leading to a different computational graph for each data sample. The training phase optimizes the learnable gating functions with an empirical loss, and the inference phase computes the appropriate sub-network through forward propagation through the gating modules~\cite{aig_veit_2017,dynamicchannelpruning_gao_2018,conditionalchannelgates_bejnordi_2019,avit_yin_2021}.

From a green AI perspective neither the dynamic or static pruning approaches are ideal. Dynamic pruning is not optimal for parallel computing due to the necessary indexing operations at inference time. Furthermore, dynamic pruning introduces overhead from the necessary connection importance computations. Static pruning can reduce computational resources at inference time, but the iterative pruning-and-fine-tuning process consumes more computational resources during the training phase. One-shot pruning after training is no better than the iterative procedure as its effectiveness heavily depends on the assumed priors, which lack verification in their validity prior to training \cite{SNIP_torr_2018}.

Our proposed pruning method forms a compressed network without the cost of significant additional training resources. It achieves this by simultaneously optimizing the network structure and parameters in a single pass. We posit that this new paradigm brings forward a significant improvement for the efficiency desiderata in pruning methods, particularly in resource-constrained environments. Particularly, the simultaneous optimization is realized with a \emph{light-weight trainable binary gating module} along with a \emph{polarising regularizer}. The polarising regularizer produces a stable sub-network that performs well for all data points towards the end of training. Inference time is reduced as the smaller static sub-network is ready-to-use. We verify the pruning scheme on two types of pruning (layer and channel) on different ResNets \cite{resnet_he_2015}, applied to three datasets with different size and number of classes (CIFAR-10, CIFAR-100, and Tiny Imagenet\cite{cifardataset_Krizhevsky_2009, tinyimagenetdataset}). Comparisons with competitive existing baselines are presented.

\subsection{Green and Red AI: Raising Awareness of the Energy Cost of AI}

\citet{greenai_schwartz_2020} were among the first authors to define the concepts of green AI (environmentally friendly AI) and red AI (heavily energy-consuming AI), suggesting that AI models should be evaluated beyond accuracy by taking into account their carbon emission and electricity usage, elapsed time, parameter count, and floating point operations (FPOs/FLOPs). \citet{carbonemissionlargemodel_patterson_2021} and \citet{carbonintensitycloud_dodge_2022} propose frameworks that quantify the carbon emission resulting from application of specific AI models on various common devices. In order to reduce the carbon emission from model training, different approaches have been commonly used. For example, model quantization can be used to reduce the elapsed time and processor memory usage \citep{modelquantization_gholami_2021}, while network distillation and network pruning approaches can be used to reduce the number of parameters and total FLOPs~\citep{knowledgedistillation_hinton_2015}.

\subsection{Network Pruning}

Network pruning aims to rank the importance of the edges in a neural network model in order to find a sub-network with the most important edges. There are generally two approaches to achieve this goal: \emph{static} or \emph{dynamic} methodologies. Static network pruning finds a unified sub-network at the end of the training, and is usually followed-up by a fine-tuning procedure to further improve the sub-network performance. This pruning scheme relies on the calculated importance scores of the edges of interest. The edge importance can be calculated, for example, by the magnitude or the influence of an edge to the final output. 

In convolutional neural networks (CNNs), pruning is usually achieved in three dimensions: depth (layers), width (channels), and resolution (feature maps) \citep{acceleratingcnns_wang_2020}. Experiments on static feature map pruning \citep{efficientconvnet_li_2016} and channel pruning \citep{channelpruning_he_2017} demonstrated a 30\% reduction in FLOPs or a 2.5$\times$ reduction in GPU time with only negligible performance degradation or even improvement in some cases. \citet{onceforall_cai_2019} expanded the problem to multi-stage pruning to make the pruning approach adaptable to different size requirements. This goal was achieved by training and fine-tuning the sub-networks with incremental size reduction while making sure the model accuracy stays the same each time the size requirement is reduced.

Dynamic pruning, on the other hand, aims to find input dependent sub-networks. Input-dependent elements are usually added to the original network to compute the importance of edges of interest. \citet{aig_veit_2017} proposed an adaptive inference graph that computes the importance of the CNN layers with a probabilistic and learnable gating module before each layer. \citet{runtimepruning_lin_2017} proposed a similar framework to prune the CNN channels by using reinforcement learning to train an optimal channel importance scoring mechanism. 

Combining both static and dynamic pruning methods can achieve synergistic gains in computational efficiency. A combined approach can benefit from the compatibility of static pruning with parallel computing, saving energy especially on GPU computation. This approach can also leverage the ability of dynamic pruning to induce networks that are input data adaptive. For example, \citet{differentiablesparsification_lee_2019} proposed a sub-differentiable sparsification method where parameters can potentially be zeroed after optimization under stochastic gradient descent. However, the non-unified sub-networks still cause excess indexing computation in parallel computing. Our work focuses on the problem of finding a unified sub-network for data following a certain distribution, by using the dynamically pruned sub-networks as intermediate states. Recent work on unifying sub-networks has involved pruning a graph representation of the neural network \citep{discoveringneuralwirings_wortsman_2019,hiddenrandomlyweighted_ramanujan_2019}

Additionally, our findings corroborate recent work on the existence of  ``lottery tickets'', \ie~pruned sub-networks that can achieve similar accuracy as the original network \citep{lottery}. To generate such networks, \citet{lottery} develop the IMP (Iterative Magnitude Pruning) scheme that involves iterative pruning and training over multiple rounds until convergence. This is different from our proposed method, which performs simultaneous training and pruning in one training session and is therefore computationally cheaper to train.

\subsection{Discrete Stochasticity and Gradient Estimation}

To obtain a stable sub-network structure through gradient-based optimization~\ie~binary activation statuses for each connection, a gating module is needed with differentiable and discrete latent variables. Discrete variables often require a relaxation or estimation due to the incompatibility of the discretisation function with back-propagation (\eg~zero gradients everywhere).

A well-known estimation approach is the Gumbel-Softmax (GS) estimator \citep{gumbelsoftmax_jang_2016}. The Gumbel-Softmax is a continuous probability distribution that can be tuned to approximate a discrete categorical distribution. The gradients \emph{w.r.t.} the categorical output distribution is well-defined for the GS distribution. This technique is often applied to generative sequence models requiring sampling under multinomial distributions \citep{gsgans_kunsner_2016,gsmultitask_shen_2021,gsnas_chang_2019}. 

An arguably simpler, but still effective approach is the straight-through estimator (STE) \citep{ste_bengio_2013}, which binarizes the stochastic output based on a threshold in the forward pass, and heuristically copies the gradient of next layer to the estimator. Experiments show that neural networks gated by the STE give the lowest error rate among other differentiable gates (multinomial and non-multinomial) \citep{ste_bengio_2013}. We describe the STE in more detail in Section \ref{ste}.

\subsection{Sparsity Regularizers}

For the purposes of parameter pruning a sparsity regularizer is often used to control the pruning ratio during training. The $\mathit{l}_1$ and $\mathit{l}_2$ regularizers are two most common. However, standard regularization functions can lead to unnecessary pruning and mistaken estimation of network connectivity importance. Regularizers that are more sensitive to network structure include $\mathit{l}_{2,0}$ and $\mathit{l}_{2,1}$ structured sparsity regularization \citep{ssr_lin_2019}, grouped on samples or on feature maps \citep{groupsparsityregularization_li_2020}~etc.

Srinivas and \citet{tristaterelu_srinivas_2015} proposed a binarizing regularizer that encourages each network connection to approach either 1 or 0 for all samples. The binarizing mechanism can also be extended to continuous activation rates. For example, \citet{polarizationreg_zhuang_2020} integrated a polarization regularizer into network pruning to force the deactivation of neurons. Networks pruned under this setting achieve the highest accuracy even at high pruning rate compared to other pruning schemes.

\section{Problem Setup: Joint Parameter and Architecture Learning}
We denote a neural network with full connectivity in the form of a graph as $\Phi:=(V, E)$, where $V$ is a set of nodes, $E$ is the set of edges $E:=\{e^{(x,y)}, \forall x,y\in V\}$. A sub-network with partial connectivity can thus be represented as $\Phi' = (V, E')$ where $ E'\subseteq E$. We also denote the transformation of a network as $f_\theta(\cdot)\equiv f(\cdot; \theta)$, where $\theta$ denotes all the parameters in a network. Each edge $e\in E$ is associated with a weight $\theta^e$. For the full network $\theta = \theta_\Phi$ , and for the sub-network $\theta=\theta_{\Phi'}$. A sub-network can be expressed in terms of a full-network using an activation matrix $\mathbf{W}_e$ with certain elements zeroed, i.e:
\begin{equation}
 \theta_{\Phi'} = \mathbf{W}_e^{\top} \theta_\Phi,
 \label{eq:prune_matrix}
\end{equation}
where $w_{e,c}\in \{0,1\}$ for every entry in the edge activation matrix $\mathbf{W}_e$ is binary, and $\theta_{\Phi'}$ is computed from a Hadamard product. In network pruning, we aim to find a sub-network $\Phi'$ and the optimal network parameters $\theta_{\Phi'}^*$ simultaneously. We estimate the optimal solution $\theta_{\Phi'}^*$ with $\hat{\theta_{\Phi'}^*}$ by minimising the empirical loss. We further reuse the settings in Eq.\ref{eq:prune_matrix} to reform the objective below: \ie
 \begin{equation}\label{eq:pareto mle2}
  \begin{aligned}
\min_{\theta_{\Phi'}, \Phi'} \mathcal{L}(f(\mathbf{x};\theta_{\Phi'}), \mathbf{y}). \\
    \min_{\theta_{\Phi}, \mathbf{W}_e} \mathcal{L}(f(\mathbf{x}; \mathbf{W}_e^\top \theta_\Phi), \mathbf{y}).
 \end{aligned}
\end{equation}

\section{Methodology}
In practice, the edge activation matrix $\mathbf{W}_e$ is not learned as a whole and each entry in the matrix is not independent. When training a sequential network the activation of earlier connections can affect the outputs of later connections, and thus also affect gradient back-propagation. A naive binary/categorical $\mathbf{W}_e$ would prevent gradients from propagating back, as a function with constant value has zero gradient. Therefore, a gradient estimator is needed as the core gating element of each connectivity. We choose the straight-through estimator (STE), as introduced in Section~\ref{ste}, as this core element.

\subsection{Network Architectures}

\begin{figure}[t]
\centering
\includegraphics[width=\columnwidth]{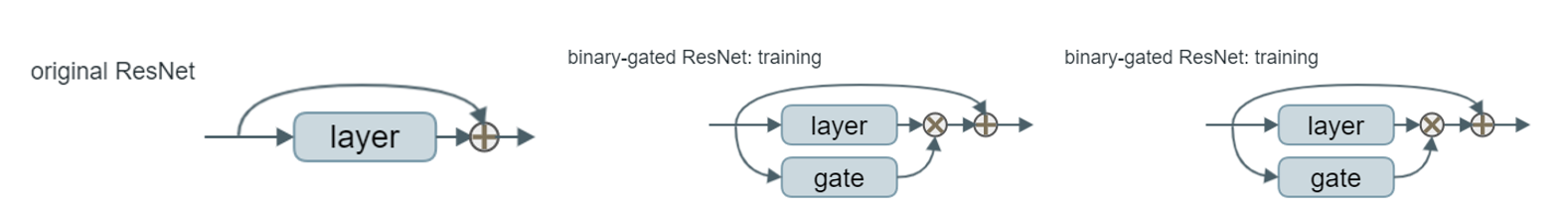}
\caption{Illustration of a gating module with binary decision as integrated into the original residual model. The learnable gating modules are trained as per other parts of the network. At inference, the gate decisions are pre-loaded, and only the network parameters whose gate decision is open are loaded and computed.}
\label{fig:overall_flow}
\end{figure}

Figure \ref{fig:overall_flow} illustrates the design for the gating module integration into a ResNet. Our pruning scheme has slightly different workflows for the training phase and the testing phase. In training, the gating modules with learnable dense layers are trained as part of the network. At inference (for validation or test), the resultant $\mathbf{W}_e$ is loaded, which decides the subset of parameters to be selected -- only the connections with a non-zero $w_{e,c}$ will be loaded for parameters and included in the forward pass. 

The choice of ResNet as the base network is based on the necessity of residual connections to avoid potential computational path termination in the middle of the network due to a deactivated full layer. For ResNet, we focus on CNN-centered layer and channel (feature map) pruning. However, we also argue that this methodology has the potential be applied to any type of connection, even in less structured pruning (\eg~selected kernel-to-kernel connections between convolutional layers) and leave empirical evidence as a direction for future work. While our method has similarities with dropout-based methods in ResNets, these involve pruning specific connections between nodes. From an architectural standpoint this doesn't necessarily reduce the number of FLOPs as there is no reduction in the number of matrix multiplications required. In contrast removing entire channels / layers has this desired effect.

\subsection{Straight-through Estimator}\label{ste}
We chose the straight-through estimator (STE) as the binary head for the gating module. The forward path of STE is a hard thresholding function:
\begin{equation}
\label{eq:ste_forward}
STE(x) = \begin{cases} 1, & \mbox{if } x > 0 \\ 
0, & \mbox{if } x\leq 0 \end{cases}.
\end{equation}

\noindent The backward gradient reflects why this estimator is known as ``straight-through'':
\begin{equation}
\label{eq:ste_backward}
\frac{\partial \mathcal L}{\partial x} = \frac{\partial \mathcal L}{\partial STE(x)}\cdot \frac{\partial STE(x)}{\partial x} = \begin{cases} \frac{\partial \mathcal L}{\partial STE(x)}, & \mbox{if } |x| \leq 1 \\ 
0, & \mbox{if } |x| > 1 \end{cases},
\end{equation}

\noindent{where the insensitive state is triggered when $|x|>1$. This is to avoid a possible scenario where a large gradient makes the STE output value stay at either 1 or 0 permanently.}
A clear advantage of the STE as the gating head is that it constitutes a lightweight module for both forward and backward propagation. In the forward pass, no other computation than a sign check is needed. In the backward pass no computation is needed. The gradient estimation, often viewed as a coarse approximation of the true gradient under noise, has been proved to positively correlate with the population gradient, and therefore gradient descent helps to minimize the empirical loss \citep{understandingSTE_yin_2019}.

\subsection{Polarising Regularization for Unified Activation}

During the dynamic pruning-style training, the matrix $\mathbf{W}_e(x)$ might not be the same for all $x \in \mathcal{X}$. To encourage a unified edge activation matrix that $\mathbf{W}_e(x) = \mathbf{W}_e(x'), \forall x,x'\in \mathcal{X}$, we introduce a polarisation regularizer $\mathcal{R}_{polar}(\{\mathbf{W}_e(x)|x\in\mathcal{X}\})$. The complete loss function is:
\begin{equation}
\mathcal L(f(\mathbf{x}),\mathbf{y}) = \mathcal L_{task}(f(\mathbf{x}),\mathbf{y}) + \lambda \mathcal{R}_{polar}(\mathbf{W}_e(\mathbf{x}))
\label{eqn:loss_func}
\end{equation}

\noindent{where $\mathcal{L}_{task}$ is the task loss,\eg~cross-entropy loss for classification tasks and mean-squared error for regression tasks, and $\lambda$ is the scale factor for polarisation regularizer.} The general form of $\mathcal{R}_{polar}(\mathbf{W}_e(\mathbf{x}))$ is in the form of an inverted parabola. Supposing $\mathbf{W}_e(x)\in \mathbb{R}^{|\mathcal{C}|}$ is flattened for all covered connections $c\in\mathcal{C}$:
\begin{equation}
    \mathcal{R}_{polar}(\mathbf{W}_e(\mathbf{x})) := \frac{1}{|\mathcal{C}|}(\mathbf{1}-\bar{\mathbf{W}}_e(\mathbf{x}))^\top \bar{\mathbf{W}}_e(\mathbf{x}),
\end{equation}

\noindent{where $\bar{\mathbf{W}}_e(\mathbf{x}) = \frac{1}{|\mathcal{X}|}\sum_{x\in\mathcal{X}}\mathbf{W}_e(x)$ is the averaged edge activation matrix over all data samples. Given the range of $\bar{\mathbf{W}}_{e,c} \in [0,1]$, this form of inverted parabola ensures that an equivalent optimum can be reached when $\bar{\mathbf{W}}_{e,c}$ reaches either boundary of the range.} Specifically, in our ResNet layer-pruning scenario, the regularisation term is written as:
\begin{equation}
    \mathcal{R}_{polar} := \frac{1}{|L|}\sum_{ly\in L} (1-\bar{g}_{ly}) \bar{g}_{ly},
\end{equation}

\noindent{where $\bar{g}_{ly} = \frac{1}{|\mathcal{X}|}\sum_{x\in\mathcal{X}}g_{ly}(x)\in [0,1]$ is the average of the gating module outputs over all input samples of the layer.} Similarly, in our ResNet channel-pruning scenario, the regularisation term is written as:
\begin{equation}
    \mathcal{R}_{polar} := \frac{1}{|L|}\sum_{ly\in L} \frac{1}{|C|}\sum_{ch\in C}(1-\bar{g}_{ly,ch}) \bar{g}_{ly,ch}
\end{equation}

\noindent{where $\bar{g}_{ly,ch} = \frac{1}{|\mathcal{X}|}\sum_{x\in\mathcal{X}}g_{ly,ch}(x)\in [0,1]$ is the average of the gating module outputs over all input samples of the channel $ch\in\mathcal{C}$ in the layer $ly\in L$.}.

\section{Experiments}\label{sec:experiments}
\subsection{Datasets and Architecture Specifications}\label{app:training_specs}
We test the effectiveness of our proposed method on the ResNet architecture \citep{resnet_he_2015} on the CIFAR-10, CIFAR-100, and Tiny Imagenet datasets, with 10, 100, and 200 classes respectively. We chose the three datasets to benchmark our work as they have widely-acknowledged test results on most variants of ResNets. Testing on both datasets shows the effectiveness of our method under both simple and complex data distributions.

Our experiments on the CIFAR datasets are conducted on one NVIDIA T4 GPU with 16GB memory. The batch size is set to 256 for CIFAR-10 and 64 for CIFAR-100. Our training is done under a staged decaying learning rate for 350 epochs (although convergence can usually be achieved before 250 epochs). The initial learning rate for both dataset is 0.1, and at each next stage will decrease to $10\%$. On CIFAR-10, the learning rate is adjusted at epochs 60, 120, and 160. On CIFAR-100, the learning rate is adjusted at epochs 125, 190, and 250. We chose stochastic gradient descent (SGD) as the optimizer, with a momentum of 0.9 and a weight decay of $5\times 10^4$. The networks and training procedures are implemented in PyTorch. When randomness is involved, we set the random seed to 1.

Our experiments on the Tiny Imagenet dataset are conducted on four NVIDIA A10 GPU with 24GB memory each. Our training is done with the SGD optimizer under a staged decaying learning rate for 1200 epochs. The learning rate is set to 0.2 initially, then decreases to $10\%$ of the previous value at 600 and 900 epochs. Our batch size is 100. In addition, we applied Puzzle mix \citep{kimICML20} for both baseline network training and pruned network training to improve the classification accuracy on Tiny Imagenet.

We apply the network to the image classification task. The pruned networks are evaluated by top-1 accuracy and FLOPs (floating-point operations). The FLOPs count is approximated by the \verb|fvcore| package \footnote{https://github.com/facebookresearch/fvcore}.

\subsection{Comparison with Existing Pruning Baselines}
We first compare the performance of our method with na\"ive baselines and other methods in the literature that are related to ours.  We follow the training specifications laid out in Section \ref{app:training_specs}. For these experiments, we choose our most performant pruning architecture based on empirical results on the test set shown later in this section, where we use the layer pruning scheme, the Gumbell softmax (section \ref{sec:gradient_estimation}) and set $\lambda_{polar} = \uparrow$ (as in Table \ref{tab:polar_activations}). Furthermore, we choose the ResNet56 as the base architecture in particular to allow for the easy comparison with other pruning schemes seen in the literature.

In addition to our original loss function, we now wish to control the pruning rate of our method. This is achieved by adding an additional sparsity regularization term in the loss function that provides a signal for the polarization regulariser to keep fewer gates open. The resulting loss function is: 
\begin{align}
\hspace{-0.05cm}
    \mathcal L(f(\mathbf{x}),\mathbf{y}) &= \mathcal L_{task}(f(\mathbf{x}),\mathbf{y}) + \lambda_{polar} \mathcal{R}_{polar}(\mathbf{W}_e(\mathbf{x})) \notag \\ & +
    \lambda_{act} \mathcal{R}_{act}(\mathbf{W}_e(\mathbf{x})) 
\end{align}
where:
\begin{equation}
    \mathcal{R}_{act}(\mathbf{W}_e(\mathbf{x})) = \frac{1}{|L|}\sum_{ly\in L} \bar{g}_{ly}
\end{equation}
is the overall average layer activation, where the average is taken across layers $l_y \in L$ and inputs $x \in \mathcal{X}$. We vary $\lambda_{act} \in [0,1]$ to change the pruning rate (\ie~the $\%\textrm{ of parameters pruned}$). In the context of our work, this is defined as:
\begin{equation}
\begin{split}
 \mathbb{E}_{x \in \mathcal{X}}
    \footnotesize{\left[\frac{\sum_{l}\#\textrm{Parameters in layer $l$: Gate $g_l(x) = 0$}}{\#\textrm{Total Parameters}} \times 100\right]}
\end{split}
\end{equation}
where the expectation over inputs accounts for the fact that some pruning schemes may not achieve perfect sub-network unification.

We consider the following na\"ive baselines, due to their empirical performance in the literature:
\begin{itemize}
    \item Na\"ive  Dropout ResNet-56: A standard classifier but with \newline $k \in \{20\%,30\%, 50\%, 60, 80\%\}$ parameters randomly pruned during testing.
    \item Na\"ive Layer Pruned ResNet-56: The same classifier but with \newline $k \in \{20\%,30\%, 50\%, 60, 80\%\}$ layer activations randomly set to $0$ during testing.
\end{itemize}

\begin{figure}[htb]
    \centering
    \includegraphics[scale=0.185]{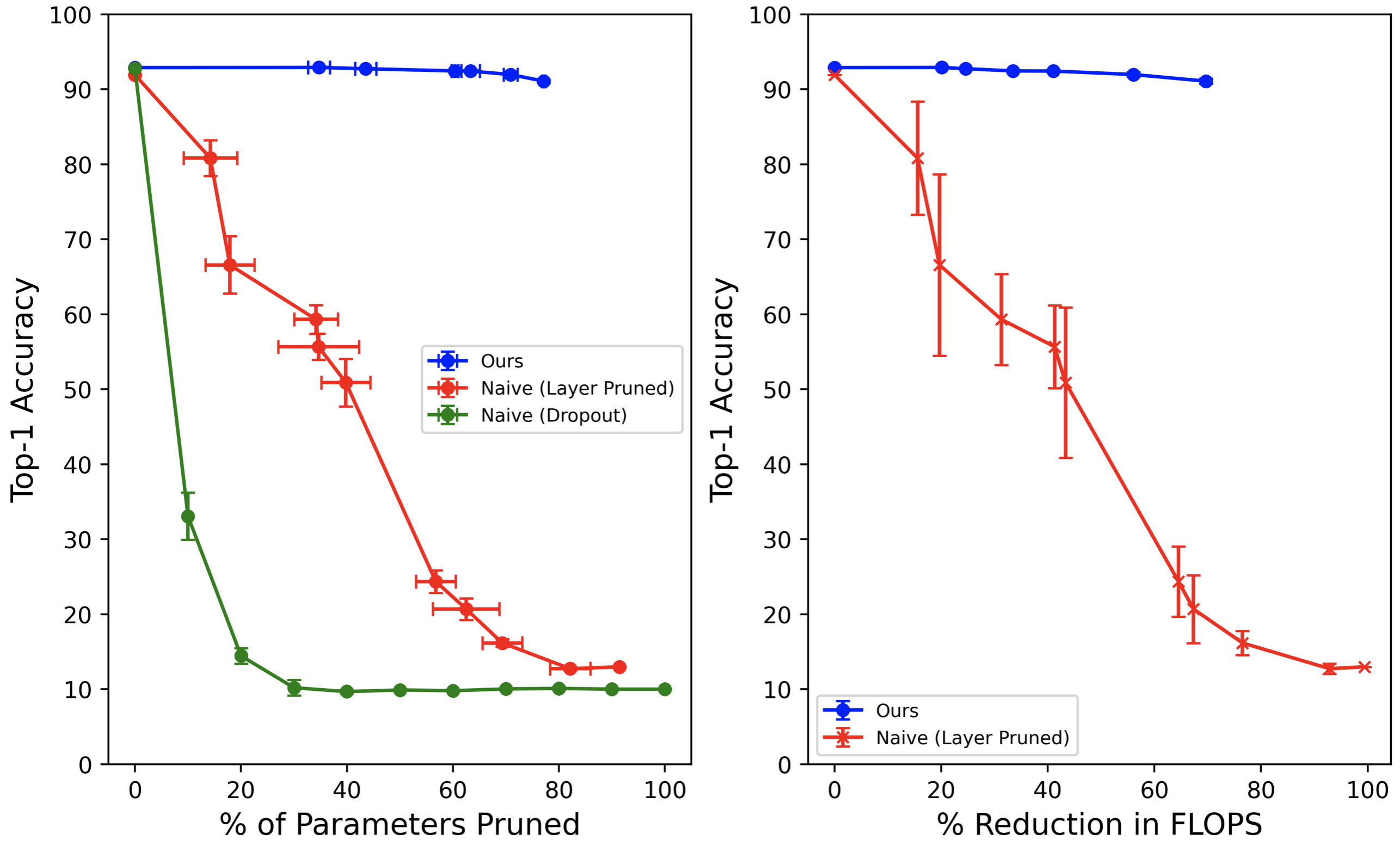}
    \caption{Comparison of our method with some na\"ive baselines on CIFAR-10 with ResNet-56. \textbf{Left}: Average pruning rate at inference vs Top-1 accuracy. \textbf{Right} \% FLOPs reduction at inference vs Top-1 accuracy. The naive dropout method does not reduce FLOPs because it still involves computation through the ``dropped'' nodes - hence the omission.}
    \label{fig:comparison_baselines}
\end{figure}
\begin{figure}[htb]
    \centering
    \includegraphics[scale=0.32]{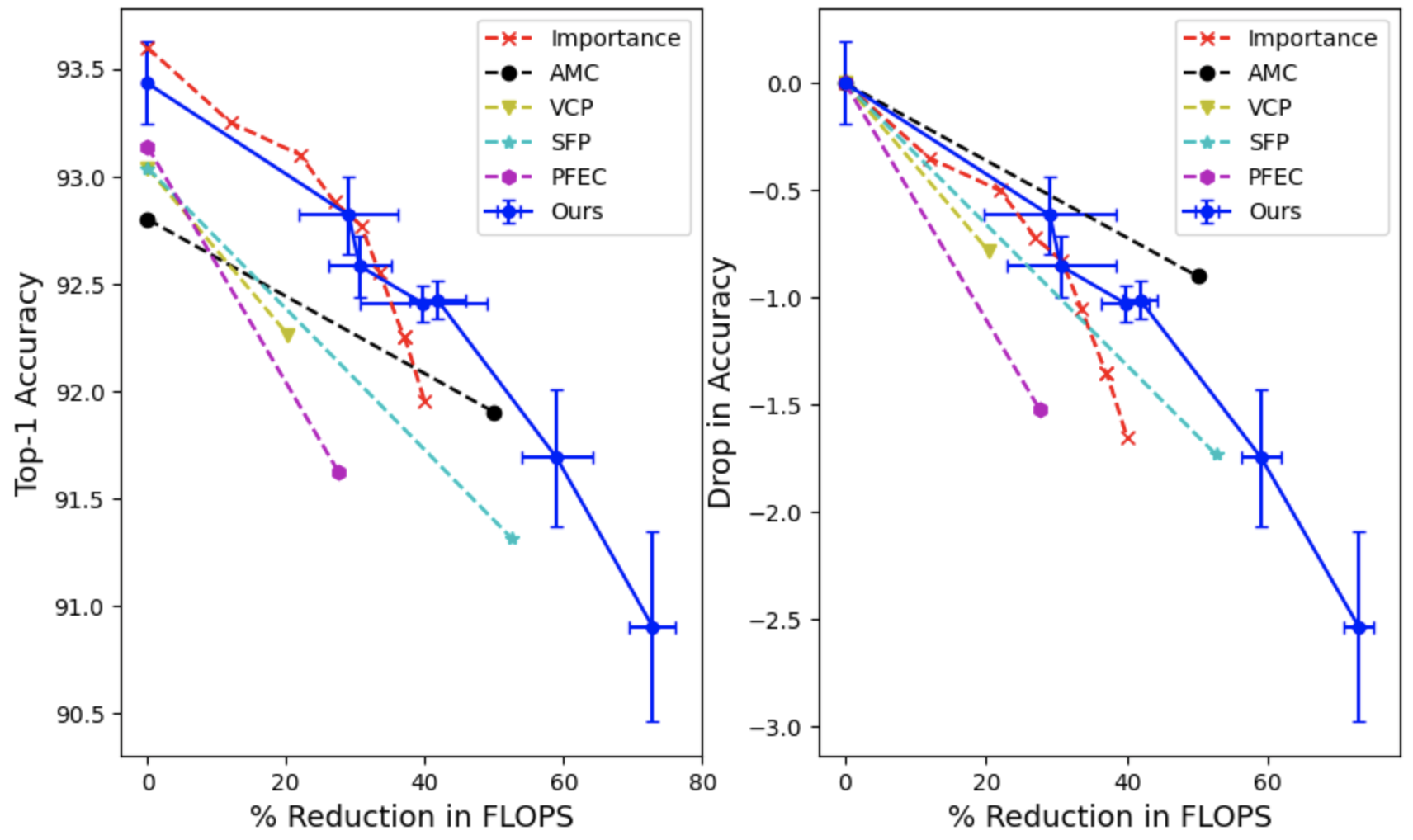}
    \caption{Comparison between our scheme and related methods in literature on CIFAR-10 with ResNet-56 at inference. \textbf{Left}: Pruning rate vs Top-1 accuracy. \textbf{Right} \% FLOPs reduction vs Top-1 accuracy drop.}
    \label{fig:comparison_lit}
\end{figure}

A visualized comparison with the na\"ive baselines is shown in Figure \ref{fig:comparison_baselines}, demonstrating our pruning method's ability to maintain network performance with a high pruning rate on the test set. 
\begin{table}
\centering
\tiny
\begin{tabular}{@{}ccccc@{}}
\toprule
\textbf{Method} & \textbf{Unpruned Accuracy} & \textbf{Pruned Accuracy} & \textbf{\% FLOPS Reduction} & \textbf{Accuracy Drop} \\ \midrule
\textbf{Ours} & 93.43 & \textbf{92.42 $\pm$ 0.14} & 41.81 $\pm$ 4.01 & \textbf{1.01 $\pm$ 0.14} \\
 AMC \citep{amc} & 92.80 & 91.90 & 50.00 & 1.10 \\
 Importance \citep{importance} & \textbf{93.60} & 91.90 & 39.90 & 1.14 \\
 SFP \citep{sfp}  & 93.59 & 92.26 & \textbf{52.60} & 1.33 \\
 CP \citep{cp}  & 92.80 & 91.80 & 50.00 & 1.00 \\
 PFEC \citep{pfec} & 93.04 & 91.31 & 27.60 & 1.73 \\
 VCP \citep{vcp}  & 93.04 & 92.26 & 20.30 & 0.78 \\ 
 \bottomrule
\end{tabular}
\caption{The performance of our method over $5$ trials against some established, related methods in the literature for $\approx 50 \%$ FLOPs reduction (\textbf{Dataset}: CIFAR-10, \textbf{Model}: ResNet-56). We note that our method offers a competitive trade-off between accuracy and FLOPs while being simple to implement. For SFP, we consider only the pre-trained variant for fair comparison as the fine-tuning variant in the paper incurs extra computational costs that are not necessarily considered.}
\label{tab:comparison}
\end{table}
Further benchmarking our method, we consider the methods that also followed the idea of simultaneous pruning and learning, as listed in Table \ref{tab:comparison}. Figure \ref{fig:comparison_lit} shows the performance of our scheme against methods found in the literature, all of which use the ResNet-56 as the base network. We note that our scheme provides competitive results not only in terms of the absolute accuracy, but also the accuracy drop resulting from pruning.

\subsection{Investigating Chanel and Layer pruning}
ResNets contain residual layers which consist of two convolutional layers and a bypasssing residual connection. 

\begin{figure}[t]
\centering
\begin{subfigure}[b]{0.2\linewidth}
         \centering
         \includegraphics[scale=0.3]{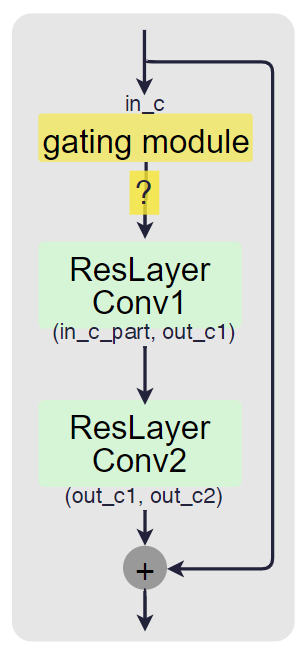}
         \caption{Channel pruning before the first convolution layer.}
         \label{fig:channel_prune_before}
\end{subfigure}
\hfill
\begin{subfigure}[b]{0.53\linewidth}
         \centering
         \includegraphics[scale=0.15]{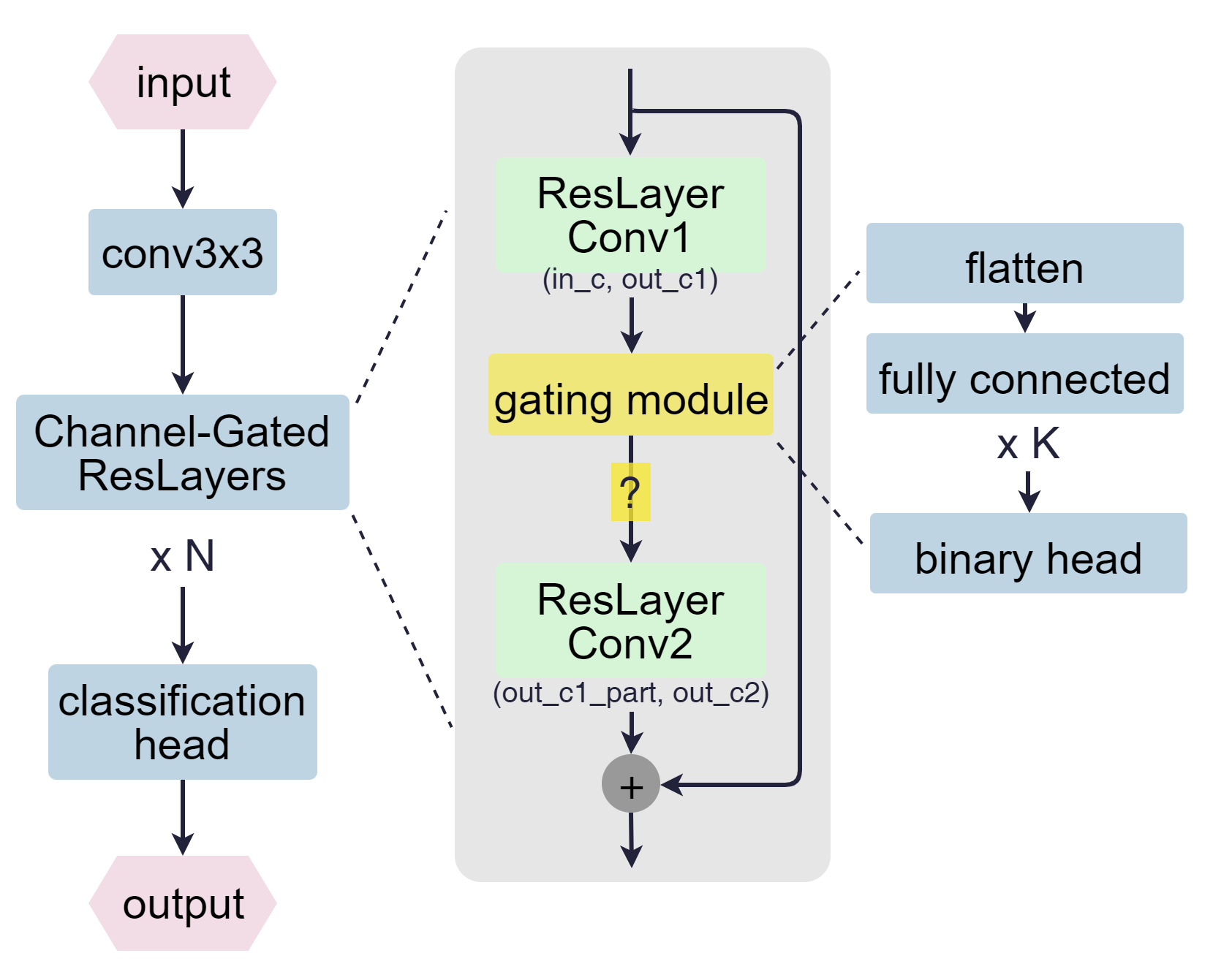}
         \caption{Channel pruning between two convolution layers.}
         \label{fig:channel_prune_middle}
\end{subfigure}
\hfill
\begin{subfigure}[b]{0.2\linewidth}
         \centering
         \includegraphics[scale=0.3]{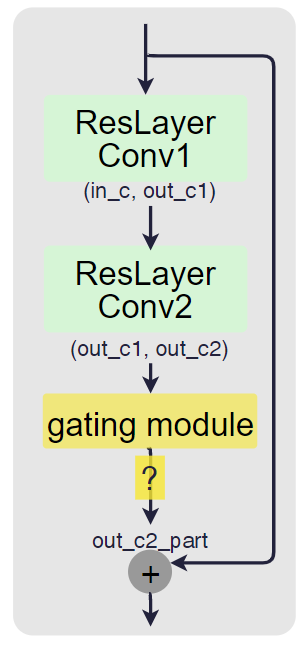}
         \caption{Channel pruning after the second convolution layer.}
         \label{fig:channel_prune_after}
\end{subfigure}
\caption{Illustration of channel-pruning gating modules in ResNet: The gating module (a) before the first convolution layer; (b) between two convolutional layers; (c) after the second convolution layer. $K{=}1$ or $2$ in our experiments.}
\label{fig:pruning_channel}
\end{figure}

In channel pruning, we experiment on two layer designs and three positions of the gating module. Specifically, we experimented with gating module architectures that consist of one and two dense layers, where Table \ref{tab:gate_module_design} shows the detailed design. For the gating module positions, we experimented on placing the gating module before the first convolutional layer, between the two convolution layers, and immediately after the second convolutional layer. Figure \ref{fig:pruning_channel} visualises the aforementioned decisions. 
Table \ref{tab:channel_structure_results_cifar10} 
shows the pruning results on CIFAR-10 under different gating module architectures and positions. For the gating module architectures (recall Table \ref{tab:gate_module_design}), results show that while all designs achieve a similar channel pruning ratio, the design with two dense layers and placed at the end of each residual layer (2FC-after) achieves the best classification accuracy that is significantly higher than most others. However, the design with 1 dense layer placed between two convolution layers (1FC-middle) also achieves a similar accuracy. Henceforth, we use our channel pruning baseline to be the \emph{2FC-after} design, as per Table \ref{tab:channel_structure_results_cifar10}, as it showed the best performance in classification accuracy and channel pruning
ratio on the test set.

\begin{table}[h]
\centering
\tiny
\begin{tabular}{c|c}
\toprule
{\textbf{Layer Gating Module}} & {\textbf{Channel Gating Module (K=2)}}\\ 
\midrule
 avg\_pool\_2d (output\_size=channel\_in) & flatten()\\
 \midrule
 dense (in\_dim=channel\_in, out\_dim=16) & dense(in\_dim=out\_channel*feature\_size, out\_dim=16)    \\
 \midrule
 batch\_norm\_1d() & batch\_norm\_1d()\\
 \midrule
 ReLU()          & ReLU() \\
 \midrule
 dense (in\_dim=16, out\_dim=1) & dense(in\_dim=16, out\_dim=1) \\
 \midrule
 STE()               & STE()   \\
\bottomrule
\end{tabular}

\caption{Layer and channel gating module design. \textbf{Left}: ``channel\_in'' is the input channel number for the first convolution layer in the residual layer. 
\textbf{Right}: ``mid\_channel'' is the output channel number for the first convolution layer, equal to the input channel number for the second convolution layer. ``feature\_size'' is the dimension of a flattened feature map. For other K values, we simply vary the dense layer number.}
\label{tab:gate_module_design}
\end{table}

\begin{table}[tb]
\small
\centering
\begin{tabular}{@{}ccc@{}}
\toprule
\textbf{Model}         & \textbf{top-1 accuracy (\%) (rel)} & \textbf{gate open ratio (\%)} \\ \midrule
baseline       & 93.68 (0)           & 100.00               \\ 
1FC-before    & 69.71 (-23.97)       & 47.82                \\ 
1FC-middle    & 90.60 (-3.08)       & 47.77                \\ 
1FC-after     & 83.89 (-9.79)       & 41.91                \\ 
2FC-before    & 85.39 (-8.29)       & 48.93                \\ 
2FC-middle    & 82.94 (-10.74)       & 49.36                \\
2FC-after     & \textbf{91.01 (-2.67)}       & 48.76                \\ 
\bottomrule
\end{tabular}

\caption{Channel pruning results under different gating module specifications on the CIFAR-10 dataset. Numbers in brackets for top-1 accuracy show the relative difference (rel) from the baseline model. ``$\{K\}$FC'' refers to $K=\{1,2\}$ dense layer(s) in gating module and ``before'', ``middle'', and ``after'' for the three gating module positions, illustrated in Figure \ref{fig:pruning_channel}.}
\label{tab:channel_structure_results_cifar10}
\end{table}

\begin{figure}[htb]
\centering
\includegraphics[width=0.55\columnwidth]{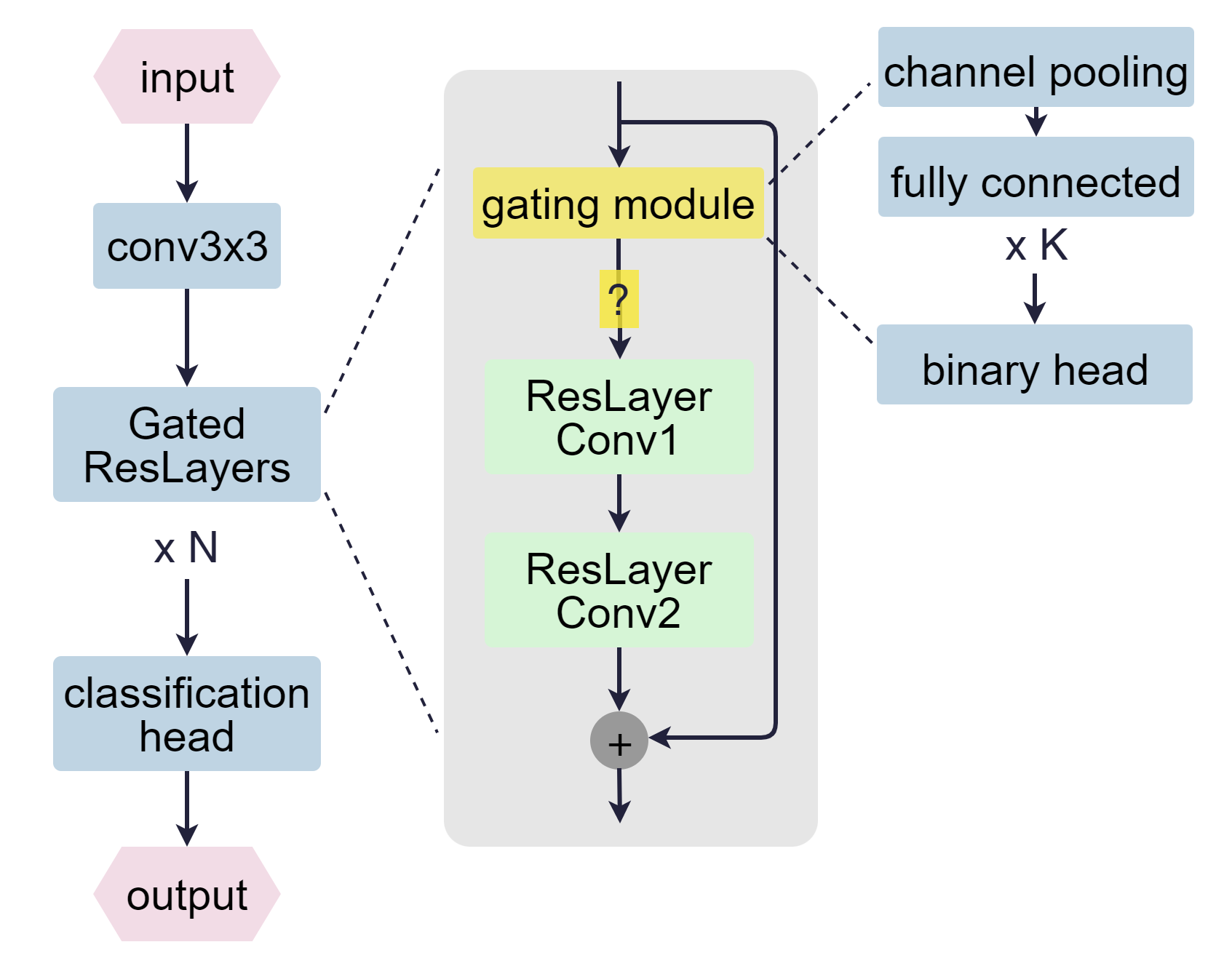}
\caption{Illustration of layer-pruning gating modules in ResNet.}
\label{fig:pruning_layer}
\end{figure}

In layer pruning, our design decisions are simplified. We add the gating module before the two convolutional layers, as seen in Figure \ref{fig:pruning_layer}, to decide whether the layer is to be computed. Table \ref{tab:gate_module_design} shows the detailed design of the gating module.

We now proceed to compare the channel and layer pruning schemes on the CIFAR-10, CIFAR-100, and Tiny Imagenet datases, as summarised in Table \ref{tab:pruning_results}. We observe that the layer pruning scheme can save at least $14\%$ of computations (FLOPs) while sacrificing accuracy of less than 2.5\%. Under the channel pruning scheme, we can save at least $22\%$ computations (FLOPs) while sacrificing accuracy of less than 3\%.

In general, we note that the layer pruned models perform better than channel pruned models (\ie~there is a lower accuracy drop) for all three datasets, even when relative differences in FLOPs are taken into account. We believe this is because under the design of ResNet, the intermediate feature maps in each residual layer is sufficiently information-compact, and any removal on the feature maps can lead to information loss. The pruning ratio of channel pruning is positive though, with an almost 50\% FLOPs reduction.

\begin{table}
\small
\centering
\begin{tabular}{@{}ccccc@{}}
\toprule
\textbf{Dataset} & \textbf{Model} & \textbf{Top-1 accuracy (\%)} & \textbf{Gate open ratio (\%)} & \textbf{FLOPs (M)(rel)} \\ \midrule
 CIFAR-10 & baseline & 93.68 (0)     & 100.00 & 255.3 (1) \\
  & layer pruned & 92.82 (-0.86) & 53.70  & 137.7 (0.54) \\
  & channel pruned & 91.01 (-2.67) & 48.76  & 189.9 (0.74) \\
  \midrule
 CIFAR-100  & baseline & 71.85 (0)     & 100.00 & 255.3 (1)   \\
   & layer pruned & 70.01 (-1.84) & 66.67  & 171.1 (0.67)\\
  & channel pruned & 66.91 (-4.94) & 51.14  & 135.41 (0.52) \\
  \midrule
  Tiny Imagenet  & baseline & 61.43 (0)     & 100.00 & 556.6 (1)   \\
   & layer pruned & 59.14 (-2.29) & 87.50  & 480.9 (0.86)\\
  & channel pruned & 58.70 (-2.73) & 66.60  & 437.84 (0.78) \\
 \bottomrule
\end{tabular}
\caption{Results of pruned networks on CIFAR-10, CIFAR-100, and Tiny Imagenet datasets. Numbers in brackets for top-1 accuracy shows the relative difference from the baseline model. FLOPs is counted in millions (M). Numbers in brackets for FLOPs shows the relative ratio from the baseline model. Baseline model is ResNet110 for CIFAR-10 and CIFAR-100. Baseline model is PreActResNet18 for Tiny Imagenet.}
\label{tab:pruning_results}
\end{table}

\subsection{Ablation Studies for Gradient Estimation}\label{sec:gradient_estimation}
We test the individual utility of the two major modules, STE and polarisation regularizer, through ablation studies. To test the utility of STE, we replace STE with sampling from Bernoulli distribution and with Gumbel-softmax. When sampling from Bernoulli distribution, we set up a threshold equal to the mean of the gating module's outputs right after the last dense layer (\ie~right before the original STE). If the output is larger than the mean, we keep the layer; otherwise we prune the layer. 

\begin{table}
\centering
\small
\begin{tabular}{@{}ccc@{}}
\toprule
\textbf{Function}     & \textbf{Top-1 accuracy (\%)} & $\mathcal{R}_{polar}$ \\
\midrule
STE       &      92.86      &  \textbf{0.00} \\
Gumbel-softmax    &   \textbf{94.05}     &  0.0039               \\
Bernoulli    &  91.96      &    0.2454    \\
\bottomrule
\end{tabular}

\caption{Results of different gating functions on CIFAR-10. $\mathcal{R}_{polar}$ taken at the end of a training session, each with the same number of epochs. A larger $\mathcal{R}_{polar}$ corresponds to less unified sub-networks after convergence. These experiments were performed on ResNet110 with $\lambda_{polar} = 3$.}
\label{tab:ablation_gating_function}
\end{table}

Table \ref{tab:ablation_gating_function} shows results from the three gating functions, experimented on CIFAR-10. We observed that other than STE, no other gating head function can result in a perfectly unified and stable sub-network. STE has utility here in terms of stabilising the evolution of the dynamic sub-networks and retaining  expected performance. However, we note that the Gumbel-softmax has a potential to achieve a better task performance while keeping a set of lightly dynamic sub-networks indicated by the low level of $\mathcal{R}_{polar}$. We leave to future work a study on how a suitable unification of the resultant dynamic sub-networks from Gumbel-softmax can be achieved to further improve the performance over STE for this task.

\begin{figure}[H]
\centering
\includegraphics[scale=0.8]{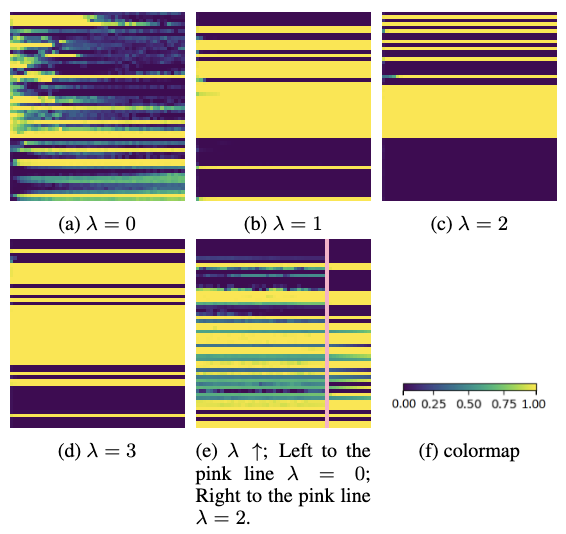}
\caption{Layer opening ratio during training under different $\lambda$. Each row represents one layer among the 54 layers. Each column represents one epoch. For $\lambda\in\{0,1,2,3\}$, we include the first 50 training epochs. For $\lambda \uparrow$, the change in $\lambda$ is separated by the pink line.}
\label{fig:polar_activations}
\end{figure}

\begin{table}
\centering
\small
\begin{tabular}{@{}ccc@{}}
\toprule
\textbf{Spec}          & \textbf{Top-1 accuracy (\%) (rel)} & \textbf{gate open ratio (\%)} \\ \midrule
baseline       & 93.68 (0)           & 100.00               \\
$\lambda_{polar}=0$    & 90.79 (-2.89) & 18.52 - 53.70  \\
$\lambda_{polar}=1$    & 92.44 (-1.24) & 53.70 \\
$\lambda_{polar}=2$     &  91.67 (-2.01)      &  \textbf{40.74}  \\
$\lambda_{polar}=3$    & 91.85 (-1.83) &  57.41  \\
$\lambda_{polar} \uparrow$ &   \textbf{93.18 (-0.50)}  & 55.56                \\
\bottomrule
\end{tabular}
\caption{Results of pruned networks under different $\lambda_{polar}$ values on CIFAR-10 (ResNet110). "$\lambda_{polar} \uparrow$" uses the settings of $\lambda_{polar}=0$ for the first 125 epochs; $\lambda=2$ for the next 65 epochs; and $\lambda=3$ for the resting epochs until end of training.}
\label{tab:polar_activations}
\end{table}

To understand the trade-off between pruning aggressiveness, accuracy retention, and computational savings, we experimented on a series of pruning aggressiveness levels, realized by tuning the hyperparamter $\lambda_{polar}$. We experimented on $\lambda_{polar}$ values varying from 0 to 3. We also tested the effect of gradually increasing regularizer weight $\lambda_{polar}$ during a training session in order to verify whether a partially trained network would affect the pruning results. Figure \ref{fig:polar_activations} shows the layer pruning evolution under different $\lambda_{polar}$ settings and Table \ref{tab:polar_activations} shows the performance of the resulting sub-networks. 

The layer pruning evolution figures show that a higher pruning aggressiveness level (a higher $\lambda_{polar}$) accelerates convergence to a unified sub-network than a lower level. On the other hand, Table \ref{tab:polar_activations} shows that a higher opening ratio does not necessarily bring more computational savings and accuracy retention. Increasing the pruning aggressiveness (introducing the penalty term) at a later stage in training, however, shows higher accuracy than other constant penalty terms, with higher computational savings than other experiments.

\section{Discussions}

\subsection{Computation Complexity}

Similar to any extra gating modules, the introduced gating module from our method adds an overhead on the overall computational complexity during training. The time complexity of STE of a single input for forward path is $\mathcal{O}(1)$ according to Eq. \ref{eq:ste_forward}. For backward path, according to Eq. \ref{eq:ste_backward}, the STE on each dimension brings only $\mathcal{O}(1)$ extra computations than the original backward path. 

Taking the setup in Table \ref{tab:gate_module_design} (left column) as an example, a gating module of $K$ fully connected layers, $d$ dimensions per layer, and $|C|$ channels as input introduces an overhead of $\mathcal{O}(dK|C|)$.

Merging the above gating module into network in Figure \ref{fig:pruning_layer}, for a network of $N$ layers and $|C|$ channels per layer, the inserted gating modules introduces an overhead of $\mathcal{O}(dNK|C|)$.

In our experiment where $K=1$ or $2$, the final overhead is shrinked to $\mathcal{O}(dN|C|)$, within which only $d$ is dependent on the gating module's design. A practically small value of $d$ is helpful in controlling the overall computational overhead.

The above overhead analysis is only applicable for training the network. After training, the optimized network parameters include the gating module's output, and a layer is either activated or deactivated at inference for arbitrary inputs. The overhead is thus excluded at inference. Our experiment (Table \ref{tab:polar_activations}) show that the pruned model can use only 55.56\% of neurons to achieve only 0.5\% of performance drop.

\section{Conclusion}

We proposed a network pruning scheme that maintains performance while being more computationally efficient. Through simultaneous parameter and structure optimization, our pruning scheme finds a stable sub-network with similar performance on the same downstream task as the full (unpruned) network. The scheme consists of a differentiable, lightweight binary gating module and regularisers to enforce data-invariance of the pruned sub-networks. Experiments on layer and channel pruning show results competitive with (or indeed better than) similar methods in literature. With fine tuning of our uncovered sub-networks, we anticipate further performance improvements - however, this is beyond the scope of our work, which aims to emphasise maximal performance gains within limited computational budgets. We look forward to testing the applicability of our pruning scheme to other base networks, datasets and tasks, hoping that our encouraging results facilitate the transition from red AI towards energy efficient and green deep learning models. 

\section{Disclaimer}
This paper was prepared for informational purposes by the Applied Innovation of AI team of JPMorgan Chase \& Co. This paper is not a product of the Research Department of JPMorgan Chase \& Co. or its affiliates. Neither JPMorgan Chase \& Co. nor any of its affiliates makes any explicit or implied representation or warranty and none of them accept any liability in connection with this paper, including, without limitation, with respect to the completeness, accuracy, or reliability of the information contained herein and the potential legal, compliance, tax, or accounting effects thereof. This document is not intended as investment research or investment advice, or as a recommendation, offer, or solicitation for the purchase or sale of any security, financial instrument, financial product or service, or to be used in any way for evaluating the merits of participating in any transaction. The described work is a prototype and is not a production deployed system.

\backmatter

\bigskip

\bibliography{sn-bibliography}


\begin{thebibliography}{51}
\ifx \bisbn   \undefined \def \bisbn  #1{ISBN #1}\fi
\ifx \binits  \undefined \def \binits#1{#1}\fi
\ifx \bauthor  \undefined \def \bauthor#1{#1}\fi
\ifx \batitle  \undefined \def \batitle#1{#1}\fi
\ifx \bjtitle  \undefined \def \bjtitle#1{#1}\fi
\ifx \bvolume  \undefined \def \bvolume#1{\textbf{#1}}\fi
\ifx \byear  \undefined \def \byear#1{#1}\fi
\ifx \bissue  \undefined \def \bissue#1{#1}\fi
\ifx \bfpage  \undefined \def \bfpage#1{#1}\fi
\ifx \blpage  \undefined \def \blpage #1{#1}\fi
\ifx \burl  \undefined \def \burl#1{\textsf{#1}}\fi
\ifx \doiurl  \undefined \def \doiurl#1{\url{https://doi.org/#1}}\fi
\ifx \betal  \undefined \def \betal{\textit{et al.}}\fi
\ifx \binstitute  \undefined \def \binstitute#1{#1}\fi
\ifx \binstitutionaled  \undefined \def \binstitutionaled#1{#1}\fi
\ifx \bctitle  \undefined \def \bctitle#1{#1}\fi
\ifx \beditor  \undefined \def \beditor#1{#1}\fi
\ifx \bpublisher  \undefined \def \bpublisher#1{#1}\fi
\ifx \bbtitle  \undefined \def \bbtitle#1{#1}\fi
\ifx \bedition  \undefined \def \bedition#1{#1}\fi
\ifx \bseriesno  \undefined \def \bseriesno#1{#1}\fi
\ifx \blocation  \undefined \def \blocation#1{#1}\fi
\ifx \bsertitle  \undefined \def \bsertitle#1{#1}\fi
\ifx \bsnm \undefined \def \bsnm#1{#1}\fi
\ifx \bsuffix \undefined \def \bsuffix#1{#1}\fi
\ifx \bparticle \undefined \def \bparticle#1{#1}\fi
\ifx \barticle \undefined \def \barticle#1{#1}\fi
\bibcommenthead
\ifx \bconfdate \undefined \def \bconfdate #1{#1}\fi
\ifx \botherref \undefined \def \botherref #1{#1}\fi
\ifx \url \undefined \def \url#1{\textsf{#1}}\fi
\ifx \bchapter \undefined \def \bchapter#1{#1}\fi
\ifx \bbook \undefined \def \bbook#1{#1}\fi
\ifx \bcomment \undefined \def \bcomment#1{#1}\fi
\ifx \oauthor \undefined \def \oauthor#1{#1}\fi
\ifx \citeauthoryear \undefined \def \citeauthoryear#1{#1}\fi
\ifx \endbibitem  \undefined \def \endbibitem {}\fi
\ifx \bconflocation  \undefined \def \bconflocation#1{#1}\fi
\ifx \arxivurl  \undefined \def \arxivurl#1{\textsf{#1}}\fi
\csname PreBibitemsHook\endcsname

\bibitem[\protect\citeauthoryear{Schwartz et~al.}{2020}]{greenai_schwartz_2020}
\begin{barticle}
\bauthor{\bsnm{Schwartz}, \binits{R.}},
\bauthor{\bsnm{Dodge}, \binits{J.}},
\bauthor{\bsnm{Smith}, \binits{N.A.}},
\bauthor{\bsnm{Etzioni}, \binits{O.}}:
\batitle{Green ai}.
\bjtitle{Commun. ACM}
\bvolume{63}(\bissue{12}),
\bfpage{54}--\blpage{63}
(\byear{2020})
\doiurl{10.1145/3381831}
\end{barticle}
\endbibitem

\bibitem[\protect\citeauthoryear{Dosovitskiy et~al.}{2020}]{ViT_dosovitskiy_2020}
\begin{botherref}
\oauthor{\bsnm{Dosovitskiy}, \binits{A.}},
\oauthor{\bsnm{Beyer}, \binits{L.}},
\oauthor{\bsnm{Kolesnikov}, \binits{A.}},
\oauthor{\bsnm{Weissenborn}, \binits{D.}},
\oauthor{\bsnm{Zhai}, \binits{X.}},
\oauthor{\bsnm{Unterthiner}, \binits{T.}},
\oauthor{\bsnm{Dehghani}, \binits{M.}},
\oauthor{\bsnm{Minderer}, \binits{M.}},
\oauthor{\bsnm{Heigold}, \binits{G.}},
\oauthor{\bsnm{Gelly}, \binits{S.}},
\oauthor{\bsnm{Uszkoreit}, \binits{J.}},
\oauthor{\bsnm{Houlsby}, \binits{N.}}:
An Image is Worth 16x16 Words: Transformers for Image Recognition at Scale.
arXiv
(2020).
\doiurl{10.48550/ARXIV.2010.11929} .
\url{https://arxiv.org/abs/2010.11929}
\end{botherref}
\endbibitem

\bibitem[\protect\citeauthoryear{Jouppi et~al.}{2020}]{tpu_jouppi_2020}
\begin{barticle}
\bauthor{\bsnm{Jouppi}, \binits{N.P.}},
\bauthor{\bsnm{Yoon}, \binits{D.H.}},
\bauthor{\bsnm{Kurian}, \binits{G.}},
\bauthor{\bsnm{Li}, \binits{S.}},
\bauthor{\bsnm{Patil}, \binits{N.}},
\bauthor{\bsnm{Laudon}, \binits{J.}},
\bauthor{\bsnm{Young}, \binits{C.}},
\bauthor{\bsnm{Patterson}, \binits{D.}}:
\batitle{A domain-specific supercomputer for training deep neural networks}.
\bjtitle{Commun. ACM}
\bvolume{63}(\bissue{7}),
\bfpage{67}--\blpage{78}
(\byear{2020})
\doiurl{10.1145/3360307}
\end{barticle}
\endbibitem

\bibitem[\protect\citeauthoryear{{EIA US}}{2021}]{energy_eia_2021}
\begin{botherref}
\oauthor{\bsnm{{EIA US}}}:
2020 Average Monthly Bill- Residential.
\url{https://www.eia.gov/electricity/sales_revenue_price/pdf/table5_a.pdf}.
Accessed: 2022-07-19
(2021)
\end{botherref}
\endbibitem

\bibitem[\protect\citeauthoryear{Denil et~al.}{2013}]{predictparams_denil_2013}
\begin{botherref}
\oauthor{\bsnm{Denil}, \binits{M.}},
\oauthor{\bsnm{Shakibi}, \binits{B.}},
\oauthor{\bsnm{Dinh}, \binits{L.}},
\oauthor{\bsnm{Ranzato}, \binits{M.}},
\oauthor{\bsnm{Freitas}, \binits{N.}}:
Predicting Parameters in Deep Learning.
arXiv
(2013).
\doiurl{10.48550/ARXIV.1306.0543} .
\url{https://arxiv.org/abs/1306.0543}
\end{botherref}
\endbibitem

\bibitem[\protect\citeauthoryear{Shafiee et~al.}{2018}]{dgate_shafiee_2018}
\begin{botherref}
\oauthor{\bsnm{Shafiee}, \binits{M.S.}},
\oauthor{\bsnm{Shafiee}, \binits{M.J.}},
\oauthor{\bsnm{Wong}, \binits{A.}}:
Dynamic Representations Toward Efficient Inference on Deep Neural Networks by Decision Gates.
arXiv
(2018).
\doiurl{10.48550/ARXIV.1811.01476} .
\url{https://arxiv.org/abs/1811.01476}
\end{botherref}
\endbibitem

\bibitem[\protect\citeauthoryear{Luo et~al.}{2019}]{thinnet_luo_2019}
\begin{barticle}
\bauthor{\bsnm{Luo}, \binits{J.}},
\bauthor{\bsnm{Zhang}, \binits{H.}},
\bauthor{\bsnm{Zhou}, \binits{H.}},
\bauthor{\bsnm{Xie}, \binits{C.}},
\bauthor{\bsnm{Wu}, \binits{J.}},
\bauthor{\bsnm{Lin}, \binits{W.}}:
\batitle{Thinet: Pruning cnn filters for a thinner net}.
\bjtitle{IEEE Transactions on Pattern Analysis `I\&' Machine Intelligence}
\bvolume{41}(\bissue{10}),
\bfpage{2525}--\blpage{2538}
(\byear{2019})
\doiurl{10.1109/TPAMI.2018.2858232}
\end{barticle}
\endbibitem

\bibitem[\protect\citeauthoryear{Cheong}{2019}]{transformerzip_cheong_2019}
\begin{bchapter}
\bauthor{\bsnm{Cheong}, \binits{R.}}:
\bctitle{transformers . zip : Compressing transformers with pruning and quantization}.
(\byear{2019})
\end{bchapter}
\endbibitem

\bibitem[\protect\citeauthoryear{Zhang and Stadie}{2019}]{oneshotpruning_zhang_2019}
\begin{botherref}
\oauthor{\bsnm{Zhang}, \binits{M.S.}},
\oauthor{\bsnm{Stadie}, \binits{B.}}:
One-Shot Pruning of Recurrent Neural Networks by Jacobian Spectrum Evaluation.
arXiv
(2019).
\doiurl{10.48550/ARXIV.1912.00120} .
\url{https://arxiv.org/abs/1912.00120}
\end{botherref}
\endbibitem

\bibitem[\protect\citeauthoryear{Lin et~al.}{2020}]{sniptransformerpruning_lin_2020}
\begin{bchapter}
\bauthor{\bsnm{Lin}, \binits{Z.}},
\bauthor{\bsnm{Liu}, \binits{J.}},
\bauthor{\bsnm{Yang}, \binits{Z.}},
\bauthor{\bsnm{Hua}, \binits{N.}},
\bauthor{\bsnm{Roth}, \binits{D.}}:
\bctitle{Pruning redundant mappings in transformer models via spectral-normalized identity prior}.
In: \bbtitle{Findings of the Association for Computational Linguistics: EMNLP 2020},
pp. \bfpage{719}--\blpage{730}.
\bpublisher{Association for Computational Linguistics},
\blocation{Online}
(\byear{2020}).
\doiurl{10.18653/v1/2020.findings-emnlp.64} .
\burl{https://aclanthology.org/2020.findings-emnlp.64}
\end{bchapter}
\endbibitem

\bibitem[\protect\citeauthoryear{Han et~al.}{2015}]{threephasepruning_han_2015}
\begin{botherref}
\oauthor{\bsnm{Han}, \binits{S.}},
\oauthor{\bsnm{Pool}, \binits{J.}},
\oauthor{\bsnm{Tran}, \binits{J.}},
\oauthor{\bsnm{Dally}, \binits{W.J.}}:
Learning both Weights and Connections for Efficient Neural Networks.
arXiv
(2015).
\doiurl{10.48550/ARXIV.1506.02626} .
\url{https://arxiv.org/abs/1506.02626}
\end{botherref}
\endbibitem

\bibitem[\protect\citeauthoryear{Zhu et~al.}{2021}]{vitpruning_zhu_2021}
\begin{botherref}
\oauthor{\bsnm{Zhu}, \binits{M.}},
\oauthor{\bsnm{Tang}, \binits{Y.}},
\oauthor{\bsnm{Han}, \binits{K.}}:
Vision Transformer Pruning.
arXiv
(2021).
\doiurl{10.48550/ARXIV.2104.08500} .
\url{https://arxiv.org/abs/2104.08500}
\end{botherref}
\endbibitem

\bibitem[\protect\citeauthoryear{Hou et~al.}{2022}]{chex_hou_2022}
\begin{botherref}
\oauthor{\bsnm{Hou}, \binits{Z.}},
\oauthor{\bsnm{Qin}, \binits{M.}},
\oauthor{\bsnm{Sun}, \binits{F.}},
\oauthor{\bsnm{Ma}, \binits{X.}},
\oauthor{\bsnm{Yuan}, \binits{K.}},
\oauthor{\bsnm{Xu}, \binits{Y.}},
\oauthor{\bsnm{Chen}, \binits{Y.-K.}},
\oauthor{\bsnm{Jin}, \binits{R.}},
\oauthor{\bsnm{Xie}, \binits{Y.}},
\oauthor{\bsnm{Kung}, \binits{S.-Y.}}:
CHEX: CHannel EXploration for CNN Model Compression.
arXiv
(2022).
\doiurl{10.48550/ARXIV.2203.15794} .
\url{https://arxiv.org/abs/2203.15794}
\end{botherref}
\endbibitem

\bibitem[\protect\citeauthoryear{Veit and Belongie}{2017}]{aig_veit_2017}
\begin{botherref}
\oauthor{\bsnm{Veit}, \binits{A.}},
\oauthor{\bsnm{Belongie}, \binits{S.}}:
Convolutional Networks with Adaptive Inference Graphs.
arXiv
(2017).
\doiurl{10.48550/ARXIV.1711.11503} .
\url{https://arxiv.org/abs/1711.11503}
\end{botherref}
\endbibitem

\bibitem[\protect\citeauthoryear{Gao et~al.}{2018}]{dynamicchannelpruning_gao_2018}
\begin{botherref}
\oauthor{\bsnm{Gao}, \binits{X.}},
\oauthor{\bsnm{Zhao}, \binits{Y.}},
\oauthor{\bsnm{Dudziak}, \binits{L.}},
\oauthor{\bsnm{Mullins}, \binits{R.}},
\oauthor{\bsnm{Xu}, \binits{C.-z.}}:
Dynamic Channel Pruning: Feature Boosting and Suppression.
arXiv
(2018).
\doiurl{10.48550/ARXIV.1810.05331} .
\url{https://arxiv.org/abs/1810.05331}
\end{botherref}
\endbibitem

\bibitem[\protect\citeauthoryear{Bejnordi et~al.}{2019}]{conditionalchannelgates_bejnordi_2019}
\begin{botherref}
\oauthor{\bsnm{Bejnordi}, \binits{B.E.}},
\oauthor{\bsnm{Blankevoort}, \binits{T.}},
\oauthor{\bsnm{Welling}, \binits{M.}}:
Batch-Shaping for Learning Conditional Channel Gated Networks.
arXiv
(2019).
\doiurl{10.48550/ARXIV.1907.06627} .
\url{https://arxiv.org/abs/1907.06627}
\end{botherref}
\endbibitem

\bibitem[\protect\citeauthoryear{Yin et~al.}{2021}]{avit_yin_2021}
\begin{botherref}
\oauthor{\bsnm{Yin}, \binits{H.}},
\oauthor{\bsnm{Vahdat}, \binits{A.}},
\oauthor{\bsnm{Alvarez}, \binits{J.}},
\oauthor{\bsnm{Mallya}, \binits{A.}},
\oauthor{\bsnm{Kautz}, \binits{J.}},
\oauthor{\bsnm{Molchanov}, \binits{P.}}:
A-ViT: Adaptive Tokens for Efficient Vision Transformer.
arXiv
(2021).
\doiurl{10.48550/ARXIV.2112.07658} .
\url{https://arxiv.org/abs/2112.07658}
\end{botherref}
\endbibitem

\bibitem[\protect\citeauthoryear{Lee et~al.}{2018}]{SNIP_torr_2018}
\begin{botherref}
\oauthor{\bsnm{Lee}, \binits{N.}},
\oauthor{\bsnm{Ajanthan}, \binits{T.}},
\oauthor{\bsnm{Torr}, \binits{P.H.S.}}:
SNIP: Single-shot Network Pruning based on Connection Sensitivity.
arXiv
(2018).
\doiurl{10.48550/ARXIV.1810.02340} .
\url{https://arxiv.org/abs/1810.02340}
\end{botherref}
\endbibitem

\bibitem[\protect\citeauthoryear{He et~al.}{2015}]{resnet_he_2015}
\begin{botherref}
\oauthor{\bsnm{He}, \binits{K.}},
\oauthor{\bsnm{Zhang}, \binits{X.}},
\oauthor{\bsnm{Ren}, \binits{S.}},
\oauthor{\bsnm{Sun}, \binits{J.}}:
Deep Residual Learning for Image Recognition.
arXiv
(2015).
\doiurl{10.48550/ARXIV.1512.03385} .
\url{https://arxiv.org/abs/1512.03385}
\end{botherref}
\endbibitem

\bibitem[\protect\citeauthoryear{Krizhevsky}{2009}]{cifardataset_Krizhevsky_2009}
\begin{bchapter}
\bauthor{\bsnm{Krizhevsky}, \binits{A.}}:
\bctitle{Learning multiple layers of features from tiny images}.
(\byear{2009}).
\burl{https://www.cs.toronto.edu/\\texttildelow kriz/learning-features-2009-TR.pdf}
\end{bchapter}
\endbibitem

\bibitem[\protect\citeauthoryear{Le and Yang}{2015}]{tinyimagenetdataset}
\begin{bchapter}
\bauthor{\bsnm{Le}, \binits{Y.}},
\bauthor{\bsnm{Yang}, \binits{X.S.}}:
\bctitle{Tiny imagenet visual recognition challenge}.
(\byear{2015}).
\burl{https://api.semanticscholar.org/CorpusID:16664790}
\end{bchapter}
\endbibitem

\bibitem[\protect\citeauthoryear{Patterson et~al.}{2021}]{carbonemissionlargemodel_patterson_2021}
\begin{botherref}
\oauthor{\bsnm{Patterson}, \binits{D.}},
\oauthor{\bsnm{Gonzalez}, \binits{J.}},
\oauthor{\bsnm{Le}, \binits{Q.}},
\oauthor{\bsnm{Liang}, \binits{C.}},
\oauthor{\bsnm{Munguia}, \binits{L.-M.}},
\oauthor{\bsnm{Rothchild}, \binits{D.}},
\oauthor{\bsnm{So}, \binits{D.}},
\oauthor{\bsnm{Texier}, \binits{M.}},
\oauthor{\bsnm{Dean}, \binits{J.}}:
Carbon Emissions and Large Neural Network Training.
arXiv
(2021).
\doiurl{10.48550/ARXIV.2104.10350} .
\url{https://arxiv.org/abs/2104.10350}
\end{botherref}
\endbibitem

\bibitem[\protect\citeauthoryear{Dodge et~al.}{2022}]{carbonintensitycloud_dodge_2022}
\begin{botherref}
\oauthor{\bsnm{Dodge}, \binits{J.}},
\oauthor{\bsnm{Prewitt}, \binits{T.}},
\oauthor{\bsnm{Combes}, \binits{R.T.D.}},
\oauthor{\bsnm{Odmark}, \binits{E.}},
\oauthor{\bsnm{Schwartz}, \binits{R.}},
\oauthor{\bsnm{Strubell}, \binits{E.}},
\oauthor{\bsnm{Luccioni}, \binits{A.S.}},
\oauthor{\bsnm{Smith}, \binits{N.A.}},
\oauthor{\bsnm{DeCario}, \binits{N.}},
\oauthor{\bsnm{Buchanan}, \binits{W.}}:
Measuring the Carbon Intensity of AI in Cloud Instances.
arXiv
(2022).
\doiurl{10.48550/ARXIV.2206.05229} .
\url{https://arxiv.org/abs/2206.05229}
\end{botherref}
\endbibitem

\bibitem[\protect\citeauthoryear{Gholami et~al.}{2021}]{modelquantization_gholami_2021}
\begin{botherref}
\oauthor{\bsnm{Gholami}, \binits{A.}},
\oauthor{\bsnm{Kim}, \binits{S.}},
\oauthor{\bsnm{Dong}, \binits{Z.}},
\oauthor{\bsnm{Yao}, \binits{Z.}},
\oauthor{\bsnm{Mahoney}, \binits{M.W.}},
\oauthor{\bsnm{Keutzer}, \binits{K.}}:
A Survey of Quantization Methods for Efficient Neural Network Inference.
arXiv
(2021).
\doiurl{10.48550/ARXIV.2103.13630} .
\url{https://arxiv.org/abs/2103.13630}
\end{botherref}
\endbibitem

\bibitem[\protect\citeauthoryear{Hinton et~al.}{2015}]{knowledgedistillation_hinton_2015}
\begin{botherref}
\oauthor{\bsnm{Hinton}, \binits{G.}},
\oauthor{\bsnm{Vinyals}, \binits{O.}},
\oauthor{\bsnm{Dean}, \binits{J.}}:
Distilling the Knowledge in a Neural Network.
arXiv
(2015).
\doiurl{10.48550/ARXIV.1503.02531} .
\url{https://arxiv.org/abs/1503.02531}
\end{botherref}
\endbibitem

\bibitem[\protect\citeauthoryear{Wang et~al.}{2020}]{acceleratingcnns_wang_2020}
\begin{botherref}
\oauthor{\bsnm{Wang}, \binits{W.}},
\oauthor{\bsnm{Chen}, \binits{M.}},
\oauthor{\bsnm{Zhao}, \binits{S.}},
\oauthor{\bsnm{Chen}, \binits{L.}},
\oauthor{\bsnm{Hu}, \binits{J.}},
\oauthor{\bsnm{Liu}, \binits{H.}},
\oauthor{\bsnm{Cai}, \binits{D.}},
\oauthor{\bsnm{He}, \binits{X.}},
\oauthor{\bsnm{Liu}, \binits{W.}}:
Accelerate CNNs from Three Dimensions: A Comprehensive Pruning Framework.
arXiv
(2020).
\doiurl{10.48550/ARXIV.2010.04879} .
\url{https://arxiv.org/abs/2010.04879}
\end{botherref}
\endbibitem

\bibitem[\protect\citeauthoryear{Li et~al.}{2016}]{efficientconvnet_li_2016}
\begin{botherref}
\oauthor{\bsnm{Li}, \binits{H.}},
\oauthor{\bsnm{Kadav}, \binits{A.}},
\oauthor{\bsnm{Durdanovic}, \binits{I.}},
\oauthor{\bsnm{Samet}, \binits{H.}},
\oauthor{\bsnm{Graf}, \binits{H.P.}}:
Pruning Filters for Efficient ConvNets.
arXiv
(2016).
\doiurl{10.48550/ARXIV.1608.08710} .
\url{https://arxiv.org/abs/1608.08710}
\end{botherref}
\endbibitem

\bibitem[\protect\citeauthoryear{He et~al.}{2017}]{channelpruning_he_2017}
\begin{botherref}
\oauthor{\bsnm{He}, \binits{Y.}},
\oauthor{\bsnm{Zhang}, \binits{X.}},
\oauthor{\bsnm{Sun}, \binits{J.}}:
Channel Pruning for Accelerating Very Deep Neural Networks.
arXiv
(2017).
\doiurl{10.48550/ARXIV.1707.06168} .
\url{https://arxiv.org/abs/1707.06168}
\end{botherref}
\endbibitem

\bibitem[\protect\citeauthoryear{Cai et~al.}{2019}]{onceforall_cai_2019}
\begin{botherref}
\oauthor{\bsnm{Cai}, \binits{H.}},
\oauthor{\bsnm{Gan}, \binits{C.}},
\oauthor{\bsnm{Wang}, \binits{T.}},
\oauthor{\bsnm{Zhang}, \binits{Z.}},
\oauthor{\bsnm{Han}, \binits{S.}}:
Once-for-All: Train One Network and Specialize it for Efficient Deployment.
arXiv
(2019).
\doiurl{10.48550/ARXIV.1908.09791} .
\url{https://arxiv.org/abs/1908.09791}
\end{botherref}
\endbibitem

\bibitem[\protect\citeauthoryear{Lin et~al.}{2017}]{runtimepruning_lin_2017}
\begin{bchapter}
\bauthor{\bsnm{Lin}, \binits{J.}},
\bauthor{\bsnm{Rao}, \binits{Y.}},
\bauthor{\bsnm{Lu}, \binits{J.}},
\bauthor{\bsnm{Zhou}, \binits{J.}}:
\bctitle{Runtime neural pruning}.
In: \beditor{\bsnm{Guyon}, \binits{I.}},
\beditor{\bsnm{Luxburg}, \binits{U.V.}},
\beditor{\bsnm{Bengio}, \binits{S.}},
\beditor{\bsnm{Wallach}, \binits{H.}},
\beditor{\bsnm{Fergus}, \binits{R.}},
\beditor{\bsnm{Vishwanathan}, \binits{S.}},
\beditor{\bsnm{Garnett}, \binits{R.}} (eds.)
\bbtitle{Advances in Neural Information Processing Systems},
vol. \bseriesno{30}.
\bpublisher{Curran Associates, Inc.}, \blocation{???}
(\byear{2017}).
\burl{https://proceedings.neurips.cc/paper/2017/file/a51fb975227d6640e4fe47854476d133-Paper.pdf}
\end{bchapter}
\endbibitem

\bibitem[\protect\citeauthoryear{Lee}{2019}]{differentiablesparsification_lee_2019}
\begin{botherref}
\oauthor{\bsnm{Lee}, \binits{Y.}}:
Differentiable Sparsification for Deep Neural Networks.
arXiv
(2019).
\doiurl{10.48550/ARXIV.1910.03201} .
\url{https://arxiv.org/abs/1910.03201}
\end{botherref}
\endbibitem

\bibitem[\protect\citeauthoryear{Wortsman et~al.}{2019}]{discoveringneuralwirings_wortsman_2019}
\begin{botherref}
\oauthor{\bsnm{Wortsman}, \binits{M.}},
\oauthor{\bsnm{Farhadi}, \binits{A.}},
\oauthor{\bsnm{Rastegari}, \binits{M.}}:
Discovering Neural Wirings.
arXiv
(2019).
\doiurl{10.48550/ARXIV.1906.00586} .
\url{https://arxiv.org/abs/1906.00586}
\end{botherref}
\endbibitem

\bibitem[\protect\citeauthoryear{Ramanujan et~al.}{2019}]{hiddenrandomlyweighted_ramanujan_2019}
\begin{botherref}
\oauthor{\bsnm{Ramanujan}, \binits{V.}},
\oauthor{\bsnm{Wortsman}, \binits{M.}},
\oauthor{\bsnm{Kembhavi}, \binits{A.}},
\oauthor{\bsnm{Farhadi}, \binits{A.}},
\oauthor{\bsnm{Rastegari}, \binits{M.}}:
What's Hidden in a Randomly Weighted Neural Network?
arXiv
(2019).
\doiurl{10.48550/ARXIV.1911.13299} .
\url{https://arxiv.org/abs/1911.13299}
\end{botherref}
\endbibitem

\bibitem[\protect\citeauthoryear{Frankle and Carbin}{2019}]{lottery}
\begin{bchapter}
\bauthor{\bsnm{Frankle}, \binits{J.}},
\bauthor{\bsnm{Carbin}, \binits{M.}}:
\bctitle{The lottery ticket hypothesis: Finding sparse, trainable neural networks}.
In: \bbtitle{7th International Conference on Learning Representations, {ICLR} 2019, New Orleans, LA, USA, May 6-9, 2019}.
\bpublisher{OpenReview.net}, \blocation{???}
(\byear{2019}).
\burl{https://openreview.net/forum?id=rJl-b3RcF7}
\end{bchapter}
\endbibitem

\bibitem[\protect\citeauthoryear{Jang et~al.}{2016}]{gumbelsoftmax_jang_2016}
\begin{botherref}
\oauthor{\bsnm{Jang}, \binits{E.}},
\oauthor{\bsnm{Gu}, \binits{S.}},
\oauthor{\bsnm{Poole}, \binits{B.}}:
Categorical Reparameterization with Gumbel-Softmax.
arXiv
(2016).
\doiurl{10.48550/ARXIV.1611.01144} .
\url{https://arxiv.org/abs/1611.01144}
\end{botherref}
\endbibitem

\bibitem[\protect\citeauthoryear{Kusner and Hernández-Lobato}{2016}]{gsgans_kunsner_2016}
\begin{botherref}
\oauthor{\bsnm{Kusner}, \binits{M.J.}},
\oauthor{\bsnm{Hernández-Lobato}, \binits{J.M.}}:
GANS for Sequences of Discrete Elements with the Gumbel-softmax Distribution.
arXiv
(2016).
\doiurl{10.48550/ARXIV.1611.04051} .
\url{https://arxiv.org/abs/1611.04051}
\end{botherref}
\endbibitem

\bibitem[\protect\citeauthoryear{Shen et~al.}{2021}]{gsmultitask_shen_2021}
\begin{bchapter}
\bauthor{\bsnm{Shen}, \binits{J.}},
\bauthor{\bsnm{Zhen}, \binits{X.}},
\bauthor{\bsnm{Worring}, \binits{M.}},
\bauthor{\bsnm{Shao}, \binits{L.}}:
\bctitle{Variational multi-task learning with gumbel-softmax priors}.
In: \beditor{\bsnm{Ranzato}, \binits{M.}},
\beditor{\bsnm{Beygelzimer}, \binits{A.}},
\beditor{\bsnm{Dauphin}, \binits{Y.}},
\beditor{\bsnm{Liang}, \binits{P.S.}},
\beditor{\bsnm{Vaughan}, \binits{J.W.}} (eds.)
\bbtitle{Advances in Neural Information Processing Systems},
vol. \bseriesno{34},
pp. \bfpage{21031}--\blpage{21042}.
\bpublisher{Curran Associates, Inc.}, \blocation{???}
(\byear{2021}).
\burl{https://proceedings.neurips.cc/paper/2021/file/afd4836712c5e77550897e25711e1d96-Paper.pdf}
\end{bchapter}
\endbibitem

\bibitem[\protect\citeauthoryear{Chang et~al.}{2019}]{gsnas_chang_2019}
\begin{botherref}
\oauthor{\bsnm{Chang}, \binits{J.}},
\oauthor{\bsnm{Zhang}, \binits{X.}},
\oauthor{\bsnm{Guo}, \binits{Y.}},
\oauthor{\bsnm{Meng}, \binits{G.}},
\oauthor{\bsnm{Xiang}, \binits{S.}},
\oauthor{\bsnm{Pan}, \binits{C.}}:
Differentiable Architecture Search with Ensemble Gumbel-Softmax.
arXiv
(2019).
\doiurl{10.48550/ARXIV.1905.01786} .
\url{https://arxiv.org/abs/1905.01786}
\end{botherref}
\endbibitem

\bibitem[\protect\citeauthoryear{Bengio et~al.}{2013}]{ste_bengio_2013}
\begin{botherref}
\oauthor{\bsnm{Bengio}, \binits{Y.}},
\oauthor{\bsnm{Léonard}, \binits{N.}},
\oauthor{\bsnm{Courville}, \binits{A.}}:
Estimating or Propagating Gradients Through Stochastic Neurons for Conditional Computation.
arXiv
(2013).
\doiurl{10.48550/ARXIV.1308.3432} .
\url{https://arxiv.org/abs/1308.3432}
\end{botherref}
\endbibitem

\bibitem[\protect\citeauthoryear{Lin et~al.}{2019}]{ssr_lin_2019}
\begin{botherref}
\oauthor{\bsnm{Lin}, \binits{S.}},
\oauthor{\bsnm{Ji}, \binits{R.}},
\oauthor{\bsnm{Li}, \binits{Y.}},
\oauthor{\bsnm{Deng}, \binits{C.}},
\oauthor{\bsnm{Li}, \binits{X.}}:
Towards Compact ConvNets via Structure-Sparsity Regularized Filter Pruning.
arXiv
(2019).
\doiurl{10.48550/ARXIV.1901.07827} .
\url{https://arxiv.org/abs/1901.07827}
\end{botherref}
\endbibitem

\bibitem[\protect\citeauthoryear{Li et~al.}{2020}]{groupsparsityregularization_li_2020}
\begin{botherref}
\oauthor{\bsnm{Li}, \binits{Y.}},
\oauthor{\bsnm{Gu}, \binits{S.}},
\oauthor{\bsnm{Mayer}, \binits{C.}},
\oauthor{\bsnm{Van~Gool}, \binits{L.}},
\oauthor{\bsnm{Timofte}, \binits{R.}}:
Group Sparsity: The Hinge Between Filter Pruning and Decomposition for Network Compression.
arXiv
(2020).
\doiurl{10.48550/ARXIV.2003.08935} .
\url{https://arxiv.org/abs/2003.08935}
\end{botherref}
\endbibitem

\bibitem[\protect\citeauthoryear{Srinivas and Babu}{2015}]{tristaterelu_srinivas_2015}
\begin{botherref}
\oauthor{\bsnm{Srinivas}, \binits{S.}},
\oauthor{\bsnm{Babu}, \binits{R.V.}}:
Learning Neural Network Architectures using Backpropagation.
arXiv
(2015).
\doiurl{10.48550/ARXIV.1511.05497} .
\url{https://arxiv.org/abs/1511.05497}
\end{botherref}
\endbibitem

\bibitem[\protect\citeauthoryear{Zhuang et~al.}{2020}]{polarizationreg_zhuang_2020}
\begin{bchapter}
\bauthor{\bsnm{Zhuang}, \binits{T.}},
\bauthor{\bsnm{Zhang}, \binits{Z.}},
\bauthor{\bsnm{Huang}, \binits{Y.}},
\bauthor{\bsnm{Zeng}, \binits{X.}},
\bauthor{\bsnm{Shuang}, \binits{K.}},
\bauthor{\bsnm{Li}, \binits{X.}}:
\bctitle{Neuron-level structured pruning using polarization regularizer}.
In: \beditor{\bsnm{Larochelle}, \binits{H.}},
\beditor{\bsnm{Ranzato}, \binits{M.}},
\beditor{\bsnm{Hadsell}, \binits{R.}},
\beditor{\bsnm{Balcan}, \binits{M.F.}},
\beditor{\bsnm{Lin}, \binits{H.}} (eds.)
\bbtitle{Advances in Neural Information Processing Systems},
vol. \bseriesno{33},
pp. \bfpage{9865}--\blpage{9877}.
\bpublisher{Curran Associates, Inc.}, \blocation{???}
(\byear{2020}).
\burl{https://proceedings.neurips.cc/paper/2020/file/703957b6dd9e3a7980e040bee50ded65-Paper.pdf}
\end{bchapter}
\endbibitem

\bibitem[\protect\citeauthoryear{Yin et~al.}{2019}]{understandingSTE_yin_2019}
\begin{botherref}
\oauthor{\bsnm{Yin}, \binits{P.}},
\oauthor{\bsnm{Lyu}, \binits{J.}},
\oauthor{\bsnm{Zhang}, \binits{S.}},
\oauthor{\bsnm{Osher}, \binits{S.}},
\oauthor{\bsnm{Qi}, \binits{Y.}},
\oauthor{\bsnm{Xin}, \binits{J.}}:
Understanding Straight-Through Estimator in Training Activation Quantized Neural Nets.
arXiv
(2019).
\doiurl{10.48550/ARXIV.1903.05662} .
\url{https://arxiv.org/abs/1903.05662}
\end{botherref}
\endbibitem

\bibitem[\protect\citeauthoryear{Kim et~al.}{2020}]{kimICML20}
\begin{bchapter}
\bauthor{\bsnm{Kim}, \binits{J.-H.}},
\bauthor{\bsnm{Choo}, \binits{W.}},
\bauthor{\bsnm{Song}, \binits{H.O.}}:
\bctitle{Puzzle mix: Exploiting saliency and local statistics for optimal mixup}.
In: \bbtitle{International Conference on Machine Learning (ICML)}
(\byear{2020})
\end{bchapter}
\endbibitem

\bibitem[\protect\citeauthoryear{He et~al.}{2018}]{amc}
\begin{bchapter}
\bauthor{\bsnm{He}, \binits{Y.}},
\bauthor{\bsnm{Lin}, \binits{J.}},
\bauthor{\bsnm{Liu}, \binits{Z.}},
\bauthor{\bsnm{Wang}, \binits{H.}},
\bauthor{\bsnm{Li}, \binits{L.-J.}},
\bauthor{\bsnm{Han}, \binits{S.}}:
\bctitle{{AMC}: {AutoML} for model compression and acceleration on mobile devices}.
In: \bbtitle{Computer Vision {\textendash} {ECCV} 2018},
pp. \bfpage{815}--\blpage{832}.
\bpublisher{Springer}, \blocation{???}
(\byear{2018}).
\doiurl{10.1007/978-3-030-01234-2_48} .
\burl{https://doi.org/10.1007/978-3-030-01234-2_48}
\end{bchapter}
\endbibitem

\bibitem[\protect\citeauthoryear{Dekhovich et~al.}{2021}]{importance}
\begin{botherref}
\oauthor{\bsnm{Dekhovich}, \binits{A.}},
\oauthor{\bsnm{Tax}, \binits{D.M.J.}},
\oauthor{\bsnm{Sluiter}, \binits{M.H.F.}},
\oauthor{\bsnm{Bessa}, \binits{M.A.}}:
Neural network relief: a pruning algorithm based on neural activity.
CoRR
\textbf{abs/2109.10795}
(2021)
{\href{https://arxiv.org/abs/2109.10795}{{2109.10795}}}
\end{botherref}
\endbibitem

\bibitem[\protect\citeauthoryear{He et~al.}{2018}]{sfp}
\begin{bchapter}
\bauthor{\bsnm{He}, \binits{Y.}},
\bauthor{\bsnm{Kang}, \binits{G.}},
\bauthor{\bsnm{Dong}, \binits{X.}},
\bauthor{\bsnm{Fu}, \binits{Y.}},
\bauthor{\bsnm{Yang}, \binits{Y.}}:
\bctitle{Soft filter pruning for accelerating deep convolutional neural networks}.
In: \bbtitle{International Joint Conference on Artificial Intelligence (IJCAI)},
pp. \bfpage{2234}--\blpage{2240}
(\byear{2018})
\end{bchapter}
\endbibitem

\bibitem[\protect\citeauthoryear{He et~al.}{2017}]{cp}
\begin{bchapter}
\bauthor{\bsnm{He}, \binits{Y.}},
\bauthor{\bsnm{Zhang}, \binits{X.}},
\bauthor{\bsnm{Sun}, \binits{J.}}:
\bctitle{Channel pruning for accelerating very deep neural networks}.
In: \bbtitle{2017 IEEE International Conference on Computer Vision (ICCV)},
pp. \bfpage{1398}--\blpage{1406}
(\byear{2017}).
\doiurl{10.1109/ICCV.2017.155}
\end{bchapter}
\endbibitem

\bibitem[\protect\citeauthoryear{Li et~al.}{2016}]{pfec}
\begin{botherref}
\oauthor{\bsnm{Li}, \binits{H.}},
\oauthor{\bsnm{Kadav}, \binits{A.}},
\oauthor{\bsnm{Durdanovic}, \binits{I.}},
\oauthor{\bsnm{Samet}, \binits{H.}},
\oauthor{\bsnm{Graf}, \binits{H.P.}}:
Pruning filters for efficient convnets.
ArXiv
\textbf{abs/1608.08710}
(2016)
\end{botherref}
\endbibitem

\bibitem[\protect\citeauthoryear{Zhao et~al.}{2019}]{vcp}
\begin{bchapter}
\bauthor{\bsnm{Zhao}, \binits{C.}},
\bauthor{\bsnm{Ni}, \binits{B.}},
\bauthor{\bsnm{Zhang}, \binits{J.}},
\bauthor{\bsnm{Zhao}, \binits{Q.}},
\bauthor{\bsnm{Zhang}, \binits{W.}},
\bauthor{\bsnm{Tian}, \binits{Q.}}:
\bctitle{Variational convolutional neural network pruning}.
In: \bbtitle{2019 IEEE/CVF Conference on Computer Vision and Pattern Recognition (CVPR)},
pp. \bfpage{2775}--\blpage{2784}
(\byear{2019}).
\doiurl{10.1109/CVPR.2019.00289}
\end{bchapter}
\endbibitem

\end{thebibliography}

\end{document}